\documentclass[10pt,twocolumn,letterpaper]{article}

\usepackage{wacv}              

\usepackage{multirow}

\definecolor{wacvblue}{rgb}{0.21,0.49,0.74}
\usepackage[pagebackref,breaklinks,colorlinks,allcolors=wacvblue]{hyperref}

\def\wacvPaperID{1307} 
\def\confName{WACV}
\def\confYear{2026}

\title{ART-ASyn: Anatomy-aware Realistic Texture-based Anomaly Synthesis Framework for Chest X-Rays}

\author{Qinyi Cao\\
The University of Sydney\\
{\tt\small qcao3447@uni.sydney.edu.au}
\and
Jianan Fan\\
The University of Sydney\\
{\tt\small jianan.fan@sydney.edu.au}
\and
Weidong Cai\\
The University of Sydney\\
{\tt\small tom.cai@sydney.edu.au}
}

\begin{document}
\maketitle
\begin{abstract}
Unsupervised anomaly detection aims to identify anomalies without pixel-level annotations. Synthetic anomaly-based methods exhibit a unique capacity to introduce controllable irregularities with known masks, enabling explicit supervision during training. However, existing methods often produce synthetic anomalies that are visually distinct from real pathological patterns and ignore anatomical structure. This paper presents a novel \textbf{A}natomy-aware \textbf{R}ealistic \textbf{T}exture-based \textbf{A}nomaly \textbf{Syn}thesis framework (\textbf{ART-ASyn}) for chest X-rays that generates realistic and anatomically consistent lung opacity related anomalies using texture-based augmentation guided by our proposed \textbf{P}rogressive \textbf{B}inary \textbf{T}hresholding \textbf{Seg}mentation method (\textbf{PBTSeg}) for lung segmentation. The generated paired samples of synthetic anomalies and their corresponding precise pixel-level anomaly mask for each normal sample enable explicit segmentation supervision. In contrast to prior work limited to one-class classification, ART-ASyn is further evaluated for zero-shot anomaly segmentation, demonstrating generalizability on an unseen dataset without target-domain annotations. Code availability is available at \url{https://github.com/angelacao-hub/ART-ASyn}.
\end{abstract}    
\section{Introduction}
\label{sec:intro}

Detecting anomalous regions in chest X-ray (CXR) images is critical for clinical decision-making, as these regions often correspond to pathologies requiring timely intervention \cite{Koyyada2023XAIChestXray, Parikh2023DLChestXray}. Beyond image-level classification, precise localization enhances diagnostic interpretability, supports downstream tasks such as treatment planning, and reduces the risk of oversight \cite{Mueller2024ChEX, DSouza2023EquiAttnCXR, Ye2024TextFreeSeg, Ye2025PrototypeSeg}. However, in CXR images, precise annotation can be expensive and time-consuming to obtain \cite{Han2022KACL, Liu2022PADChest}. Hence, detecting anomalous regions without requiring pixel-level ground-truth annotation is highly desirable, enabling scalable and cost-effective deployment of anomaly detection models in clinical settings \cite{Fan2024SeeingUnseen}.

Unsupervised anomaly detection (UAD) has emerged as a promising solution, aiming to learn the distribution of normal data and identify anomalies as deviations from this learned distribution without requiring annotated anomalies or pixel-level supervision \cite{Lagogiannis2024SurveyUPD, Xiang2023SQUID, Xiang2024ChestConsistency}. Reconstruction-based methods include both image and feature reconstruction are widely used for anomaly detection, typically rely on generative models such as autoencoders (AEs) \cite{Baur2019DAEBrainMR, Zimmerer2019VAEAnomaly, Zimmerer2019CeVAE, Baur2021Autoencoders, Fan2022HHVAE, Zhou2023SCVAE}, generative adversarial networks (GANs) \cite{Schlegl2017AnoGAN, Schlegl2019fAnoGAN}, or diffusers \cite{Wolleb2022DiffusionAD, Wyatt2022AnoDDPM, Beizaee2025DeCoDiff}. These models are trained exclusively on normal images with the objective of reconstructing them accurately, under the assumption that unseen abnormal regions will not be faithfully reconstructed and thus exhibit high reconstruction errors during inference. However, these methods suffer from two notable limitations. First, they are prone to the “identical shortcut” problem \cite{You2022UniAD, You2022ADTR}, where the models simply learn to denoise or replicate the input image regardless of whether the content is normal or abnormal. Second, they face a trade-off between sensitivity and precision \cite{Bercea2023MorphAEus}, increasing the level of distortion can amplify the reconstruction error in normal regions, while insufficient distortion can lead to poor modeling of normal appearance, thereby reducing the reliability of the anomaly score.

In contrast to reconstruction-based approaches, synthesizing-based methods generate artificial anomalies by manipulating normal images, offering the advantage of known spatial locations for explicit supervision. For instance, CutPaste \cite{Li2021CutPaste} relocates patches within the same image to introduce anomalies, while NSA \cite{Schlueter2022NSA} employs Poisson image editing and parameter tuning to create more natural-looking synthetic anomalies. Despite these advances, such methods often produce visually distinct but anatomically implausible anomalies, neglecting domain-specific structural regularities. Building on this idea, AnatPaste \cite{Sato2023AnatPaste} incorporates threshold-based lung segmentation to constrain anomaly synthesis within lung regions. However, its reliance on a single threshold and a simplistic cut-and-paste strategy fails to account for the appearance of more delicate body structures. Consequently, models trained with such data may overfit to superficial cues and fail to generalize to real-world anomalies \cite{Li2024CropMixPaste}.

To overcome these limitations, we propose an \textbf{A}natomy-aware \textbf{R}ealistic \textbf{T}exture-based \textbf{A}nomaly \textbf{Syn}thesis method (\textbf{ART-ASyn}) for CXRs which synthesizes realistic lung opacity related anomalies using texture brushes that mimic the appearance of chest vessels, airways, and body tissues under CXRs, producing pathologies that resemble real cases. To ensure anatomical consistency, we introduce a \textbf{P}rogressive \textbf{B}inary \textbf{T}hresholding \textbf{Seg}mentation method (\textbf{PBTSeg}) that progressively refines thresholded masks to accurately delineate lung regions and constrain anomaly placement within anatomically plausible areas. To reduce reconstruction errors in normal regions and guide the model to focus on abnormal areas, ART-ASyn leverages pixel-level anomaly masks derived by binarizing the synthesized anomalies, enabling explicit supervision for anomaly segmentation and localized loss functions that operate only within abnormal regions. Building on prior studies on CXRs that focus solely on one-class classification, we extend evaluation to anomaly segmentation by testing our model in a zero-shot setting, trained on CheXpert \cite{Irvin2019CheXpert} and ZhangLab \cite{Kermany2018ZhangLab}, and evaluated on the unseen QaTa-COV19-v2 \cite{Yamac2021QaTaCOV19v2} dataset without any pixel-level annotations from the target domain. Our key contributions are as follows:

\begin{itemize}
\item We propose \textbf{ART-ASyn}, a novel anomaly synthesis method that generates realistic CXR lung opacity related anomalies guided by \textbf{PBTSeg}, a progressive binary thresholding lung segmentation method to ensure anatomical consistency.
\item The ART-ASyn framework \textbf{reduces reconstruction errors in normal regions} and guides the model to focus on abnormal areas by utilizing pixel-level anomaly masks for explicit anomaly segmentation supervision and localized reconstruction loss.
\item We demonstrate the model’s generalizability through zero-shot anomaly segmentation on unseen dataset, without using target-domain annotations.
\end{itemize}

\begin{figure*}
  \centering
  \includegraphics[width=0.7\linewidth]{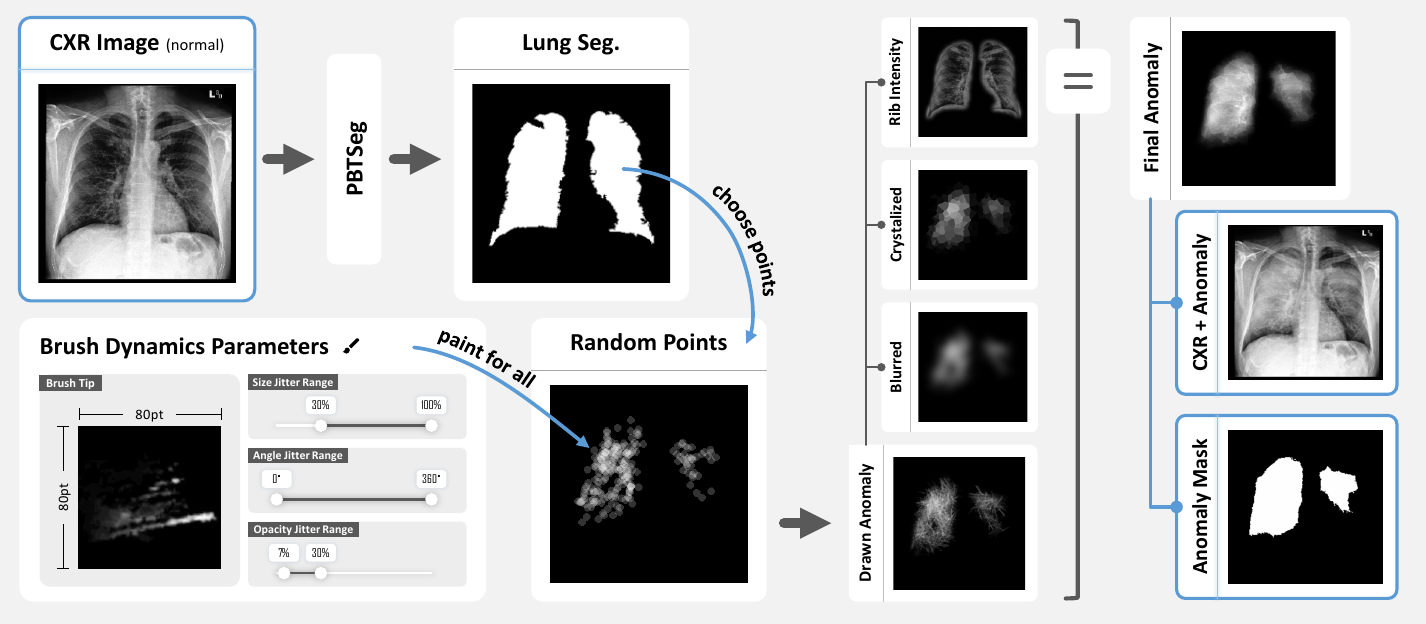}

   \caption{Overview of the ART-ASyn synthesis pipeline for generating realistic synthetic lung opacity related anomalies in CXR images. Random points within the lung region segmented by PBTSeg are painted using a dynamic brush to form an initial anomaly. A series of transformations are then combined and applied to this anomaly, which is overlaid onto the original CXR image to create a synthetic abnormal image.}
   \label{fig:ART-ASyn}
\end{figure*}
\section{Related Work}
\label{sec:related_work}

For unsupervised anomaly detection (UAD), the most closely related approaches fall into two main categories: reconstruction-based and synthesis-based methods. In addition, there exist \emph{embedding-based} methods \cite{Konz2023Unsupervised,Ruff2018Deep,Roth2022PatchCore} that frame anomaly detection as an out-of-distribution (OOD) problem. However, these methods are primarily designed for image-level classification and lack the ability to provide fine-grained localization of anomalies, so we do not discuss them in detail here.

\subsection{Reconstruction-based Methods}
Reconstruction-based methods assume that models trained exclusively on normal data will produce higher reconstruction errors when presented with abnormal regions. These methods typically follow a corrupt-and-reconstruct paradigm, where the input is deliberately corrupted and then reconstructed to align with the learned distribution of normal data. While reconstruction-based methods have good generalizability to adapt across different imaging modalities, they struggle to distinguish between fine-grained anatomical structures and pathological variations, leading to poor localization of subtle anomalies.

Traditional autoencoder-based models \cite{Baur2019DAEBrainMR, Zimmerer2019VAEAnomaly} and GAN-based methods \cite{Schlegl2017AnoGAN, Schlegl2019fAnoGAN} laid the foundation for this paradigm but tend to overgeneralize, suffering from the “identical shortcut” issue where abnormal regions remain unreconstructed and appear abnormal, leading to false negatives. With the rise of diffusion models in image synthesis \cite{Dhariwal2021DiffusionBeatsGANs}, they were adapted for anomaly detection due to their stability and fine-grained generation. Wolleb \etal \cite{Wolleb2022DiffusionAD} first applied diffusion to medical anomaly detection, AnoDDPM \cite{Wyatt2022AnoDDPM} introduced simplex noise to improve anomaly perturbation, and AutoDDPM \cite{Bercea2023AutoDDPM} proposed iterative masking and resampling to enhance reconstruction granularity.

More recent methods have improved fidelity and localization through architectural or training refinements. For example, \textbf{IterMask2} \cite{Liang2024IterMask2} introduced an iterative mask refinement strategy guided by high-frequency Fourier components to enhance reconstruction fidelity in unsupervised brain lesion segmentation, reaching the state of the art on brain MRI lesion segmentation. \textbf{UniAS} \cite{Ma2025UniAS} adopted unified feature-level reconstruction frameworks for multi-class segmentation, combining CNN or Transformer-CNN decoders to reconstruct semantically rich representations and improve localization. Meanwhile, \textbf{DeCo-Diff} \cite{Beizaee2025DeCoDiff} reformulated the diffusion process to selectively correct abnormal regions in latent space while preserving normal areas, achieving state-of-the-art results on natural image benchmarks.

\subsection{Synthesizing-based Methods}
Synthesizing-based anomaly detection simulates realistic irregularities within normal images to train models in a self-supervised manner for detecting or localizing anomalies without requiring annotated abnormal data. These methods offer the advantage of precisely controlled anomaly regions that enable targeted supervision, but often produce unrealistic appearances that fail to capture true anatomical variability, limiting their generalization to real pathological cases.

FPI \cite{Tan2022FPI} inserts a foreign patch from a different image into a normal one and trains the model to regress the interpolation factor at the pixel-level, thereby highlighting anomalous regions. PII \cite{Tan2021PII} improves upon FPI by employing Poisson blending to seamlessly integrate the patch into its new context, yielding more realistic and less visually detectable synthetic anomalies. Concurrently, CutPaste \cite{Li2021CutPaste} introduces a simpler yet effective strategy in which rectangular patches are cut and pasted within the same image and trained to classify between normal and altered images, enhancing sensitivity to spatial irregularities. NSA \cite{Schlueter2022NSA} extends PII by rescaling and shifting source patches before pasting, while enforcing anatomical plausibility through object-aware constraints. Building on this, AnatPaste \cite{Sato2023AnatPaste} incorporates anatomical priors by restricting pasted patches to segmented lung regions, using simple thresholding and morphological operations for mask generation. Most recently, CropMixPaste \cite{Li2024CropMixPaste} integrates multi-scale cropping, MixUp blending, and blurring to simulate local density variations in lung CT images, coupled with an attentive refinement block to guide the model through a self-supervised proxy task.

\section{Method}
\label{sec:method}

\begin{figure*}
  \centering
  \includegraphics[width=0.95\linewidth]{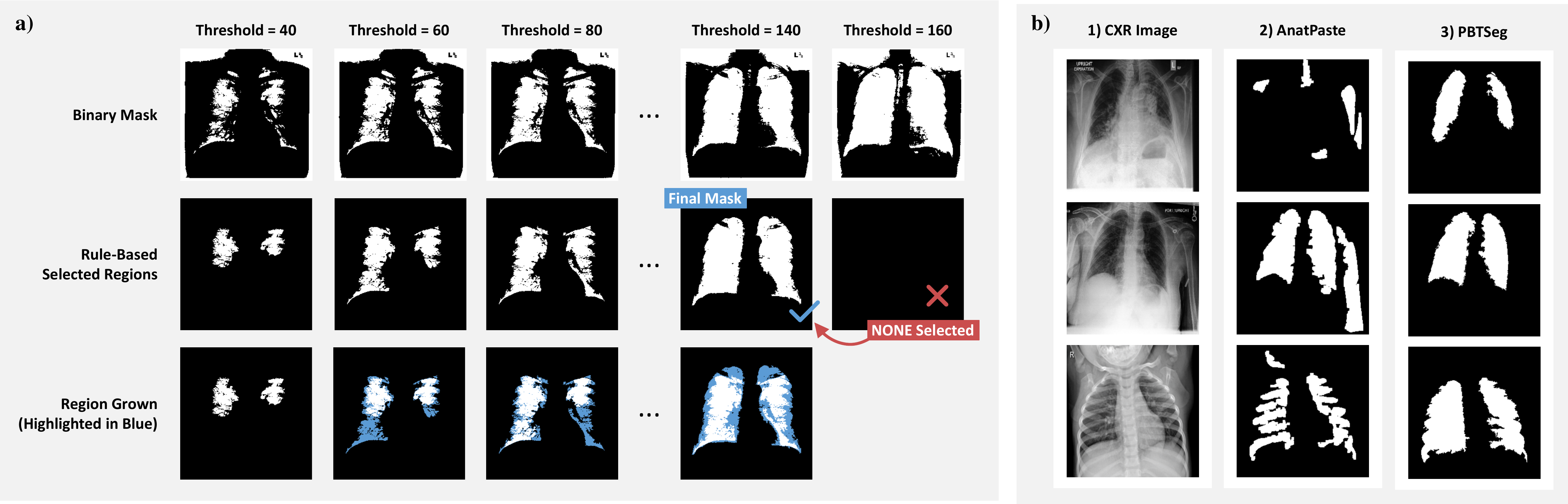}

   \caption{(a) PBTSeg process illustration of lung mask segmentation using multi-thresholding, where the second row shows candidate regions after rule-based selection for each threshold, and the last row illustrates the iterative process of updating the segmentation regions; (b) Comparison of lung segmentation results using our PBTSeg method against the threshold-based approach adopted in AnatPaste for CXR images.}
   \label{fig:PBTSeg}
\end{figure*}

In this section, we present our ART-ASyn framework, designed for chest X-rays. The core of the framework comprises two stages: (1) PBTSeg, as shown in \cref{fig:PBTSeg}, which provides accurate lung masks for anatomically valid anomaly placement, and (2) ART-ASyn, as shown in \cref{fig:ART-ASyn}, generates paired samples consisting of synthetic lung opacity related anomalies and precise pixel-level anomaly masks. These paired samples enable explicit supervision for anomaly segmentation, while the design of the training framework encourages the encoder to focus on learning normal representations. We detail each component below.

\subsection{PBTSeg}
Accurate lung segmentation is critical for synthesizing anatomically plausible anomalies in chest X-rays. While supervised lung segmentation techniques can produce precise masks, they contradict the fundamental goal of unsupervised anomaly detection, which seeks to eliminate the reliance on pixel-level annotations. A previous method \cite{Sato2023AnatPaste} attempted to use a single global intensity threshold to segment the lung regions, leveraging the fact that lungs generally appear as low-intensity areas in CXRs. However, such an approach is limited by its sensitivity to intensity variation across patients and imaging conditions. Low thresholds tend to produce fragmented and incomplete masks, while high thresholds often result in the inclusion of non-lung structures such as the mediastinum or external artifacts. To overcome these limitations while maintaining the unsupervised setting, we propose PBTSeg, a rule-based segmentation method that progressively explores multiple thresholds to identify the most representative lung regions.

Given a chest X-ray image $I_{\text{norm}}$, we define a series of ascending thresholds $T = \{t_1, t_2, \dots, t_n\}$, where $t_1 < t_2 < \dots < t_n$. For each threshold $t_i \in T$, a binary mask $B_i$ is computed as:
\begin{equation}
    B_i(x, y) = 
        \begin{cases}
        1, & \text{if } I(x, y) < t_i \\
        0, & \text{otherwise}
    \end{cases}.
\end{equation}

Connected components within $B_i$ are then identified, denoted as a set $\mathcal{C}_i = \{C_{i1}, C_{i2}, \dots, C_{im}\}$, where each $C_{ij} \subseteq B_i$ is a maximal set of spatially connected foreground pixels. Each connected region $C_{ij}$ is evaluated based on a set of geometric and positional criteria, including approximate roundness, aspect ratio (height greater than width), limited contact with image borders, and an area constraint to filter out small noise and large non-lung structures. From the valid candidates in $\mathcal{C}_i$, we select the two largest regions, $\mathcal{C}_i^{(1)}$ and $\mathcal{C}_i^{(2)}$, and assign them to the left and right lung masks based on their corresponding half of the image.

This process is applied iteratively across all thresholds, progressively updating the masks when a candidate region provides a more complete lung representation. The final left and right lung masks, denoted as $M_{\text{left}}$ and $M_{\text{right}}$, respectively, capture the largest plausible lung areas without including external artifacts. The final combined lung mask is then defined as $M_{\text{lung}} = M_{\text{left}} \cup M_{\text{right}}$. Examples of segmentation results are illustrated in \cref{fig:PBTSeg}b.

\subsection{ART-ASyn Framework}
\subsubsection{Synthesis Method}
To generate anatomically realistic and visually diverse lung opacity related anomalies, we propose a multi-stage texture synthesis pipeline guided by the lung segmentation mask produced by PBTSeg, see \cref{fig:ART-ASyn}. Given a normal CXR image $I_{\text{norm}}$, we first extract the lung mask $M_{\text{lung}}$ and randomly sample points within the lung region to serve as anchors for anomaly generation. A dynamic brush tool is used to paint textures around the sampled points, where each brush stroke is defined by controllable parameters such as angle jitter, size jitter, and opacity jitter. These strokes form the base of our synthetic abnormality $A_{\text{base}}$, enabling spatial variability and local intensity fluctuations similar to those observed in real lesions.

To enhance realism, we apply a series of transformations to the initial painted anomaly $A_{\text{base}}$:
\begin{enumerate}
    \item \textbf{Crystallization} ($T_{\text{cryst}}$) transformation to introduce irregular and sharp-edged structures, stimulating irregular margins often seen in malignant pleural effusion and enhancing the visual boundary contrast of the lesion.
    \item \textbf{Blurring} ($T_{\text{blur}}$) transformation to emulate dense tissue or the gradual base of consolidations.
    \item \textbf{Intensity scaling} ($T_{\text{rib}}$) transformation to highlight the high-density bony structures by scaling the anomaly according to the rib intensity map of $I_{\text{norm}}$, thereby preserving radiographic realism and ensuring that synthetic anomalies do not occlude the consistently high-intensity rib shadows.
\end{enumerate}
The conceptual description of the transformation combination is illustrated in the middle section of \cref{fig:ART-ASyn}, forming the final anomaly $A_{\text{final}}$. The detailed explanation of the transformations and their respective composition is provided in Supplementary Material (\S1). The corresponding synthetic anomaly image $I_{\text{syn}}$ is obtained by adding $A_{\text{final}}$ to $I_{\text{norm}}$, and the binary anomaly mask $M_{\text{anomaly}}$ is simultaneously generated by thresholding $A_{\text{final}}$ as illustrated in the right section of \cref{fig:ART-ASyn}.

This design ensures that synthetic pathologies not only preserve anatomical consistency but also reflect the diversity and uncertainty of real abnormalities, representing patterns that resemble airways, vessels, or soft tissue disruptions commonly seen in CXRs.

\subsubsection{Training and Inference}
The ART-ASyn framework is trained using sample triplets $\{I_{\text{norm}}, I_{\text{syn}}, M_{\text{anomaly}}\}$, where $I_{\text{norm}}$ is a normal CXR image, $I_{\text{syn}}$ is the corresponding synthetic anomaly image of $I_{\text{norm}}$, and $M_{\text{anomaly}}$ is the binary anomaly mask of $I_{\text{syn}}$. 
The framework adopts a U-Net architecture following \cite{Liang2024IterMask2}, which consists of an encoder-decoder structure.

During training, we pass the synthetic anomaly image $I_{\text{syn}}$ to the encoder $f_{\text{enc}}(\cdot)$ and encourage its feature representation to match that of $I_{\text{norm}}$. This is regulated by a feature-alignment loss $\mathcal{L}_{\text{feat}}$ to enforce anomaly-invariant encoding:
\begin{equation}
\mathcal{L}_{\text{feat}} = 1 - \cos(f_{\text{enc}}(I_{\text{syn}}), f_{\text{enc}}(I_{\text{norm}})).
\end{equation}
The decoder $f_{\text{dec}}(\cdot)$ then reconstructs the image $\hat{I}$ from the encoded feature representation $f_{\text{enc}}(I_{\text{syn}})$ to closely mimic that of $I_{\text{norm}}$. Reconstruction quality is measured using two losses. The global reconstruction loss is computed over the entire image:
\begin{equation}
\mathcal{L}_{\text{global}} = \| I_{\text{norm}} - \hat{I} \|^2.
\end{equation}
The local reconstruction loss focuses on the anomaly region:
\begin{equation}
\mathcal{L}_{\text{local}} = \frac{\| (I_{\text{norm}} - \hat{I}) \cdot M_{\text{anomaly}} \|^2}{\sum M_{\text{anomaly}} + \epsilon}.
\end{equation}
We also provide explicit anomaly segmentation guidance by incorporating a Dice loss between the predicted and ground-truth anomaly masks. The predicted mask $\hat{M}_{\text{anomaly}}$ is obtained by applying a threshold $\tau$ to the pixel-wise reconstruction error between the synthetic input $I_{\text{syn}}$ and its reconstruction $\hat{I}$:
\begin{equation}
\hat{M}_{\text{anomaly}} = (I_{\text{syn}}-\hat{I})^2 \geq \tau,
\end{equation}
and the corresponding Dice loss is computed by:
\begin{equation}
\mathcal{L}_{\text{dice}} = \text{Dice}(\hat{M}_{\text{anomaly}}, M_{\text{anomaly}}).
\end{equation}
An analysis of the sensitivity of the binarization threshold $\tau$ is provided in Supplementary Material (\S2).
The overall loss combines all components:
\begin{equation}
\mathcal{L}_{\text{total}} = \mathcal{L}_{\text{local}} + \mathcal{L}_{\text{global}} + \mathcal{L}_{\text{feat}} + \mathcal{L}_{\text{dice}}.
\end{equation}
The independent contribution of each loss term is provided in Supplementary Material (\S3).

At inference time, given an unseen image $I$, the model reconstructs it as $\hat{I}$. Anomaly maps are obtained by pixel-wise absolute differences $A = |I - \hat{I}|$, and binary anomaly masks are produced by thresholding $A$ using the same rule as in training.

\section{Experiment Setup}
\label{sec:experiment_setup}

\subsection{Datasets}

We evaluate ART-ASyn across three chest X-ray datasets with different data sources and supervision levels.
\begin{itemize}
    \item\textbf{CheXpert} \cite{Irvin2019CheXpert} comprises a training set of 223,414 CXR images and a test set of 500 CXR images with expert-verified labels. From this dataset, we use the 12,234 healthy frontal-view images from the training set for model training and evaluate on the original test set for image-level anomaly detection.
    \item\textbf{ZhangLab} \cite{Kermany2018ZhangLab} comprises 3,150 frontal-view CXRs with binary labels (normal or abnormal), where 1,349 images are healthy and the rest are unhealthy. We use the entire healthy images from the training set, and evaluate on the original test set for image-level anomaly detection.
    \item\textbf{QaTa-ZeroShot} is a custom zero-shot anomaly segmentation benchmark built on the QaTa-COV19-v2 dataset \cite{Yamac2021QaTaCOV19v2}, a COVID-19 CXR collection with pixel-level lung abnormality annotations. We train on healthy images from CheXpert and ZhangLab, and evaluate on all 9,250 samples from QaTa-COV19-v2, including both its original training and test splits to maximize the diversity and volume of test cases. This setup avoids using target-domain pixel annotations, allowing fair evaluation of zero-shot anomaly segmentation.
\end{itemize}

\subsection{Evaluation Metrics}

We employ different metrics for different datasets based on the granularity of available supervision.
\textbf{(1) Image-level evaluation} is conducted on the classification datasets CheXpert and ZhangLab to assess the model’s ability to distinguish between normal and abnormal cases. It is performed using Area Under the Receiver Operating Characteristic curve (AUC):
\begin{equation}
\text{AUC} = \int_{0}^{1} TPR(FPR^{-1}(x)) dx,
\end{equation}
where $TPR$ and $FPR$ denote true and false positive rates and Average Precision (AP):
\begin{equation}
\text{AP} = \sum_n (R_n - R_{n-1}) P_n,
\end{equation}
where $P_n$, $R_n$ are precision and recall at the $n$-th threshold. These metrics evaluate the model’s ranking ability across thresholds and are widely adopted in anomaly detection benchmarks. \textbf{(2) Pixel-level evaluation} is conducted on the QaTa-ZeroShot dataset utlizing the Dice similarity coefficient to quantify segmentation overlap:
\begin{equation}
\text{Dice} = \frac{2|M_{\text{pred}} \cap M_{\text{gt}}|}{|M_{\text{pred}}| + |M_{\text{gt}}|}.
\end{equation}
This metric directly assesses the spatial accuracy of predicted anomaly masks and is critical for evaluating lesion localization.

\begin{table*}[!t]
  \centering
  \begin{tabular}{@{}llccccc@{}}
    \toprule
    \multirow{2}{*}{Category} & \multirow{2}{*}{Model} & \multicolumn{2}{c}{CheXpert} & \multicolumn{2}{c}{ZhangLab} & \multicolumn{1}{c}{QaTa-ZeroShot} \\
    \cmidrule(lr){3-4} \cmidrule(lr){5-6} \cmidrule(l){7-7}
    & & AP & AUC & AP & AUC & Dice \\
    \midrule
    \multirow{3}{*}{\shortstack[l]{Reconstruction-\\Based}} 
      & IterMask2 (MICCAI 2024) & 0.7209 & 0.5205 & 0.6784 & 0.5914 & 0.1652 \\
      & DeCo-Diff (CVPR 2025) & 0.6201 & 0.3669 & 0.5241 & 0.3651 & 0.1505 \\
      & UniAS (WACV 2025)           & \underline{0.7926} & 0.6640 & 0.7907 & 0.6638 & 0.0030 \\
    \midrule
    \multirow{3}{*}{\shortstack[l]{Synthesis-\\Based}} 
      & CutPaste (CVPR 2021)        & 0.7527 & 0.6207 & 0.8451 & 0.8076 & 0.0017 \\
      & NSA (ECCV 2022)             & 0.7156 & 0.4829 & 0.7938 & 0.6506 & \underline{0.2371} \\
      & AnatPaste (iScience 2023)   & 0.7704 & \textbf{0.7825} & \underline{0.8848} & \underline{0.9103} & 0.1244 \\
    \midrule
    \textbf{Ours} & \textbf{ART-ASyn} & \textbf{0.8639} & \underline{0.7125} & \textbf{0.9621} & \textbf{0.9350} & \textbf{0.3275} \\
    \bottomrule
  \end{tabular}
  \caption{Quantitative comparison of anomaly detection models across image-level (CheXpert, ZhangLab) and pixel-level (QaTa-ZeroShot) benchmarks. The best results are highlighted in bold text and the second best results are underlined.}
  \label{tab:quant_results}
\end{table*}

\begin{table*}[!t]
  \centering
  \begin{tabular}{@{}llccccc@{}}
    \toprule
    \multirow{2}{*}{Training Setting} & \multirow{2}{*}{Synthesis Method} & \multicolumn{2}{c}{CheXpert} & \multicolumn{2}{c}{ZhangLab} & \multicolumn{1}{c}{QaTa-ZeroShot} \\
    \cmidrule(lr){3-4} \cmidrule(lr){5-6} \cmidrule(l){7-7}
    & & AP & AUC & AP & AUC & Dice \\
    \midrule
    \multirow{5}{*}{U-Net} 
    & None & 0.7394 & 0.5929 & 0.7938 & 0.6842 & 0.0707 \\
    & CutPaste & 0.7491 & 0.5805 & \underline{0.7615} & 0.6313 & 0.0934 \\
    & NSA & 0.7292 & 0.5334 & 0.7578 & \underline{0.7039} & \underline{0.1959} \\
    & AnatPaste & \underline{0.7781} & \underline{0.6028} & 0.6652 & 0.5807 & 0.0254 \\
    & \textbf{Ours (ART-ASyn)}         & \textbf{0.8250} & \textbf{0.6671} & \textbf{0.8124} & \textbf{0.7457} & \textbf{0.2771} \\
    \midrule
    \multirow{4}{*}{U-Net + Loss} 
    & CutPaste & 0.7886	& 0.6420 & 0.8060 & 0.6581 & 0.0937 \\
    & NSA & 0.7849 & 0.6334 & 0.8172 & \underline{0.7037} & \underline{0.2125} \\
    & AnatPaste & \underline{0.7908} & \underline{0.6793} & \underline{0.8182} & 0.7576 & 0.0524 \\
    & \textbf{Ours (ART-ASyn)}         & \textbf{0.8639} & \textbf{0.7125} & \textbf{0.9621} & \textbf{0.9350} & \textbf{0.3275} \\
    \bottomrule
  \end{tabular}
  \caption{Ablation of synthetic anomaly generation methods under our baseline model under two training settings: standard global reconstruction loss (top) and our custom loss (bottom). Bold indicates the best result, underline indicates the second best for each training setting.}
  \label{tab:ablation}
\end{table*}


\subsection{Baseline Methods}

We compare ART-ASyn with state-of-the-art baselines from both reconstruction-based and synthesis-based anomaly detection paradigms. Each baseline is selected to reflect either strong performance, architectural diversity, or alignment with our problem setting.

For the reconstruction-based baselines, we selected three recent state-of-the-art methods that illustrate different reconstruction paradigms. IterMask2 \cite{Liang2024IterMask2} and DeCo-Diff \cite{Beizaee2025DeCoDiff} are image-level approaches that restore normal anatomy and detect anomalies from reconstruction residuals. In contrast, UniAS \cite{Ma2025UniAS} operates at the feature-level, learning compact latent representations to directly predict anomaly scores and masks without full image reconstruction. Among them, IterMask2 is particularly relevant as a direct architectural baseline for our framework, owing to its U-Net backbone and focus on anomaly localization in medical images.

For synthesis-based baselines, we selected three representative methods: CutPaste \cite{Li2021CutPaste}, which introduces a simple patch-replacement strategy as a foundational synthesis baseline; NSA \cite{Schlueter2022NSA}, which employs Poisson blending and parameter tuning to generate handcrafted but more realistic anomalies; and AnatPaste \cite{Sato2023AnatPaste} advances synthesis by explicitly constraining the placement of pasted patches within lung fields, thereby introducing anatomical priors, making it the closest prior method to ours.

We omit older autoencoder baselines (e.g., VAE, AE, f-AnoGAN) due to their well-documented underperformance on UAD benchmarks \cite{Bercea2023MorphAEus, Ma2025UniAS} and include only competitive, peer-reviewed baselines from recent literature.

\section{Result and Discussion}
\label{sec:result&discussion}

\begin{figure*}
  \centering
  \includegraphics[width=1\linewidth]{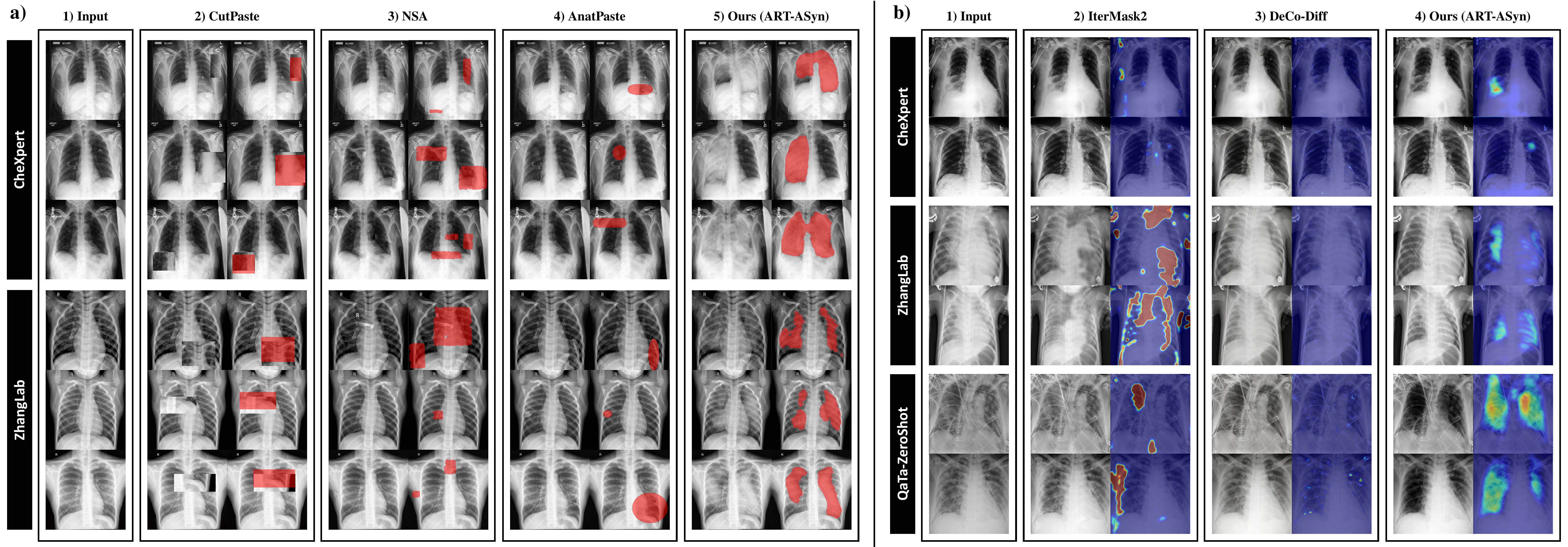}

   \caption{(a) Comparison of anomaly synthesis methods, for each method, the first column illustrates the anomaly synthesized and the second column highlights the anomaly area in red; (b) Comparison of image-level reconstruction, for each method, the first column illustrates the reconstructed image and the second column shows the pixel-wise difference $A$ in heatmap format.}
   \label{fig:compare}
\end{figure*}

\begin{figure*}
  \centering
  \includegraphics[width=0.7\linewidth]{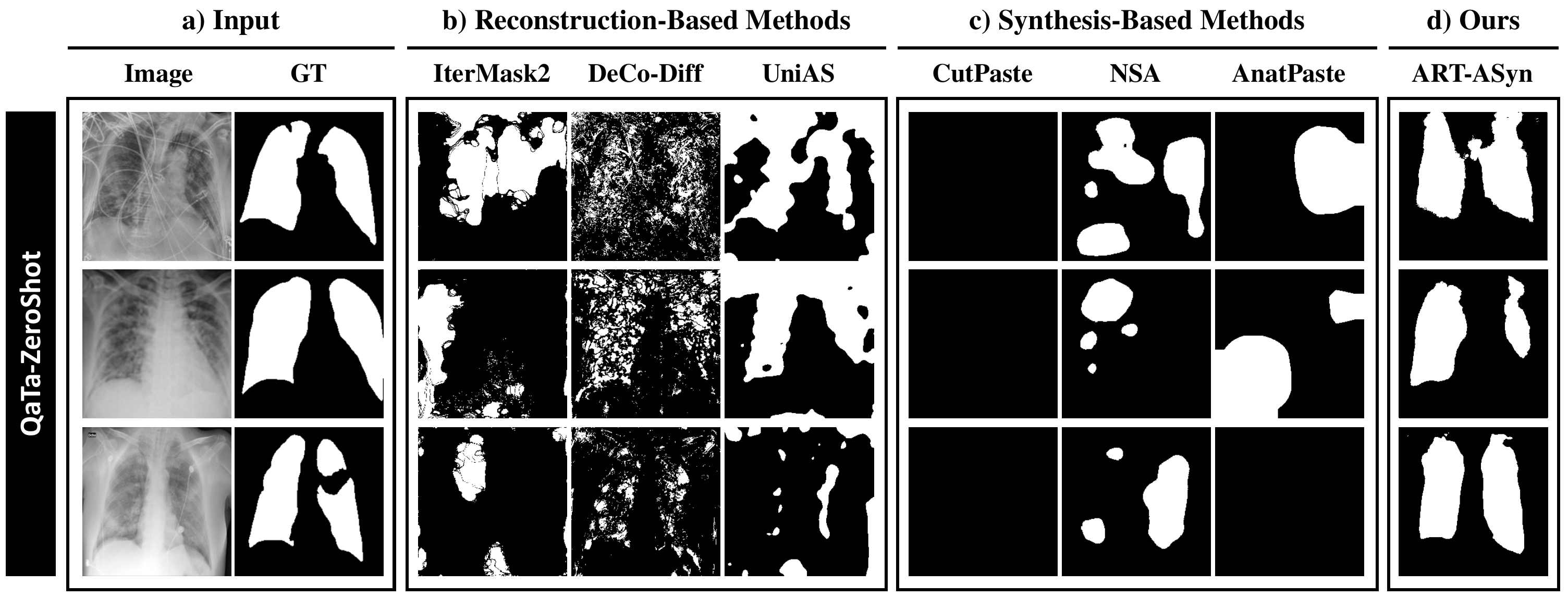}

   \caption{Comparison of anomaly segmentation on the QaTa-ZeroShot dataset.}
   \label{fig:all_methods}
\end{figure*}

In this section, we present both quantitative and qualitative results to validate the effectiveness of ART-ASyn on chest X-ray anomaly detection and segmentation tasks. Our method is evaluated on three datasets: CheXpert, ZhangLab, and QaTa-ZeroShot. As shown in \cref{tab:quant_results}, ART-ASyn outperforms existing reconstruction-based and synthesis-based baselines on most evaluation metrics.

\subsection{Quantitative Results}

\subsubsection{Image-level Anomaly Detection}
On the CheXpert dataset, ART-ASyn achieves an AP of 0.8639 and an AUC of 0.7125. It surpasses the second-best method in AP, UniAS, which achieves 0.7926, marking an improvement of 0.0713. While ART-ASyn achieves consistent gains on most metrics, its AUC on CheXpert is lower than AnatPaste's score of 0.7825. This discrepancy may be attributed to the base model architecture, as further discussed in the ablation study.

On the ZhangLab dataset, ART-ASyn obtains an AP of 0.9621 and an AUC of 0.9350, outperforming the second-best model, AnatPaste, which achieves 0.8848 and 0.9103 respectively, with improvements of 0.0773 in AP and 0.0247 in AUC. These results indicate that while existing methods already perform strongly on image-level classification, ART-ASyn delivers notable gains in precision and overall ranking performance. The performance across two independently sourced datasets demonstrates that the model generalizes well to unseen data distributions.

\subsubsection{Pixel-level Anomaly Segmentation}
In the more challenging zero-shot pixel-level segmentation task on QaTa-ZeroShot, ART-ASyn achieves a Dice score of 0.3275, significantly surpassing the second-best baseline NSA with a Dice score of 0.2371 by 0.0904. Other methods such as UniAS and CutPaste, with Dice scores of 0.003 and 0.0017 respectively, show almost no meaningful segmentation capability, despite their high performance in image-level tasks. This discrepancy reveals a critical limitation of models that rely solely on global classification signals or unstructured anomaly synthesis. In contrast, ART-ASyn benefits from pixel-level anomaly masks during training and anatomically constrained synthesis via PBTSeg, enabling the model to focus explicitly on the reconstruction and localization of pathological regions. The results highlight the effectiveness of our design in producing coherent and clinically meaningful anomaly masks under a zero-shot setting.

\subsection{Ablation Study}

To further validate the contribution of our proposed synthesis method, we perform an ablation study by integrating alternative synthesizing-based anomaly generation strategies of CutPaste, AnatPaste, and NSA into our U-Net model backbone and evaluating them both with and without our proposed loss functions (reconstruction, feature alignment, and segmentation loss). The results are summarized in \cref{tab:ablation}. Under identical model and training conditions, ART-ASyn consistently outperforms all other synthesis strategies.

When using the basic U-Net model without any anomaly synthesis (i.e., trained only on $I_{\text{norm}}$) and standard global reconstruction loss, the performance is moderate on CheXpert and ZhangLab, with AP/AUC of 0.7394/0.5929 and 0.7938/0.6842 respectively. On QaTa-ZeroShot, however, the performance is notably limited, achieving only 0.0707 in Dice. After adding synthesis methods, we observe that their image-level performance (AP/AUC) remains comparable to their original implementations reported in \cref{tab:quant_results}. For instance, NSA achieves 0.7156/0.4829 in AP/AUC under its original setting, and 0.7292/0.5334 when re-implemented under our U-Net baseline. However, in most cases, AUC tends to drop slightly when these methods are used with U-Net. For example, CutPaste achieves 0.6207 AUC in its original setting but drops to 0.5805 under the U-Net structure. This performance degradation can be attributed to the architectural design of U-Net, which is tailored for dense prediction tasks and emphasizes fine-grained spatial details rather than global feature separation required for optimal image-level ranking. As a result, while AP remains high, AUC is adversely affected due to weaker global discriminative power.

When trained with our proposed loss design, all synthesis-based methods exhibit improved performance across all metrics. For instance, CutPaste under U-Net improves from AP/AUC of 0.7491/0.5805 to 0.7886/0.6420 with our loss, exceeding its original AP 0.7527 by 0.0359. This improvement validates the effectiveness of our loss design in leveraging both pixel-level anomaly masks and normal feature alignment to enhance both segmentation quality and anomaly discrimination.

In summary, this ablation demonstrates that both components, our anatomically guided synthesis and the proposed loss functions are critical to achieving high performance. While alternative synthesis methods can benefit from our loss design, ART-ASyn still achieves the best performance, highlighting the importance of anatomical realism and paired feature supervision in synthetic anomaly generation.

\subsection{Qualitative Results}

We provide qualitative comparisons across three aspects: lung segmentation, synthetic anomaly visualization and result visualization to visually demonstrate the effectiveness of our proposed ART-ASyn framework.

\subsubsection{Lung Segmentation Comparison}

We compare the anatomical accuracy of our PBTSeg-generated lung masks against the threshold-based method used in AnatPaste. As shown in \cref{fig:PBTSeg}b, PBTSeg produces more complete and artifact-free lung boundaries, especially in cases with low contrast or overlapping soft tissues, which is crucial for anatomically constrained anomaly synthesis.

\subsubsection{Synthetic Anomaly Visualization}

To illustrate the realism and variability of synthetic anomalies, we present visual comparisons of anomalies generated by CutPaste, NSA, AnatPaste, and our proposed ART-ASyn in \cref{fig:compare}a. CutPaste produces visually abrupt and unrealistic anomalies with sharp, artificial boundaries that do not resemble true pathological patterns. In contrast, NSA and AnatPaste generate anomalies that are overly subtle and barely distinguishable to the naked eye, and in some cases, anomalies appear outside anatomically plausible regions. ART-ASyn, however, produces lesions that better capture the shape, texture, and intensity distribution of real abnormalities, including soft edges, irregular boundaries, and radiographic consistency with ribs and surrounding tissues.

\subsubsection{Image-Level Reconstruction Visualization}

We visualize image-level reconstructions in \cref{fig:compare}b to highlight the behavior of different models when reconstructing abnormal CXRs. ART-ASyn successfully preserves normal anatomical structures while suppressing abnormal regions, resulting in clearer residual anomaly maps. In contrast, IterMask2 often reconstructs normal regions as abnormal, introducing false residuals that reduce interpretability. DeCo-Diff, on the other hand, produces overly smooth reconstructions where almost no noticeable differences remain, making it difficult to localize anomalies.

\subsubsection{Anomaly Segmentation Visualization}

In \cref{fig:all_methods}, we present a comprehensive visualization of anomaly segmentation outputs from all baseline methods alongside ART-ASyn. Compared to other approaches, ART-ASyn generates more localized and interpretable segmentation masks that closely align with the true lesion locations. In contrast, other methods either fail to detect the abnormal regions or incorrectly classify large portions of normal anatomy as abnormal.
\section{Conclusion}
\label{sec:conclusion}

In this work, we proposed ART-ASyn, an anatomy-aware texture-based anomaly synthesis framework for unsupervised anomaly detection and segmentation in chest X-rays. Our method synthesizes lung opacity related anomalies through texture augmentation and spatial transformations, guided by Progressive Binary Thresholding Segmentation (PBTSeg) to ensure anatomically valid placement. The framework enables fine-grained anomaly localization via explicit segmentation supervision and feature-aligned training. Experiments on three benchmarks show that ART-ASyn outperforms both reconstruction-based and synthesis-based baselines across image-level and pixel-level metrics, and achieves promising generalization in zero-shot segmentation. A current limitation is that PBTSeg is approximately 10 times slower compared to simple thresholding, though this represents only a one-time preprocessing cost. Moreover, our scope is restricted to lung opacity related abnormalities in chest X-rays, and extending to broader pathologies remains future work.

\end{document}